\documentclass[fleqn,10pt]{wlscirep}
\usepackage[utf8]{inputenc}
\usepackage[T1]{fontenc}
\usepackage{multirow}
\usepackage{makecell}
\usepackage{ifdraft}
\usepackage{graphicx}
\setkeys{Gin}{draft=false} 
\usepackage{multicol, multirow}
\usepackage{amsmath, amssymb}
\usepackage{cmap}
\usepackage{adjustbox}
\usepackage[T1]{fontenc}
\usepackage[obeyDraft]{todonotes}
\usepackage{multirow}
\usepackage{soul}
\usepackage{tabularx,rotating,booktabs,multirow}

\newcommand{\bmx}[0]{\begin{bmatrix}}
\newcommand{\emx}[0]{\end{bmatrix}}

\newcommand{\vect}[1]{\mathbf{#1}}

\newcommand{\matr}[1]{\mathbf{#1}}

\newcommand{\va}[0]{\vect{a}}

\newcommand{\vc}[0]{\vect{c}}

\newcommand{\vh}[0]{\vect{h}}

\newcommand{\vz}[0]{\vect{z}}

\newcommand{\vs}[0]{\vect{s}}
\newcommand{\vf}[0]{\vect{f}}
\newcommand{\vi}[0]{\vect{i}}
\newcommand{\vo}[0]{\vect{o}}

\newcommand{\vg}[0]{\vect{g}}

\newcommand{\vL}[0]{\vect{L}}

\newcommand{\vE}[0]{\matr{E}}

\newcommand{\RR}[0]{\mathbb{R}}

\newcommand{\sigmoid}{\sigma}

\newcolumntype{L}[1]{>{\raggedright\let\newline\\\arraybackslash\hspace{0pt}}m{#1}}
\newcolumntype{C}[1]{>{\centering\let\newline\\\arraybackslash\hspace{0pt}}m{#1}}
\newcolumntype{R}[1]{>{\raggedleft\let\newline\\\arraybackslash\hspace{0pt}}m{#1}}

\title{Medical Image Captioning via Generative Pretrained Transformers}

\author[1,$\dagger$]{Alexander Selivanov} 
\author[2,$\dagger$]{Oleg Y. Rogov}
\author[3]{Daniil Chesakov}
\author[3]{Artem Shelmanov}
\author[1]{Irina Fedulova}
\author[2,*]{Dmitry V. Dylov}
\affil[1]{Alexander Selivanov and Irina Fedulova are with the Philips Innovation Labs Rus, Skolkovo Technopark 42 building 1 Bolshoi boulevard,  Moscow, Russia, 121205.}
\affil[2]{Oleg Y. Rogov and Dmitry V. Dylov* are with the Skolkovo Institute of Science and Technology, Bolshoy blvd., 30/1, Moscow 121205, Russia}
\affil[3]{Daniil Chesakov and Artem Shelmanov are with the Sber AI Lab, Kutuzovsky Ave, 32 bld. 1, Moscow, 121170.}
\affil[$\dagger$]{Contributed equally}
\affil[*]{Corresponding author e-mail: d.dylov@skoltech.ru}

\begin{abstract}
The automatic clinical caption generation problem is referred to as proposed model combining the analysis of frontal chest X-Ray scans with structured patient information from the radiology records. We combine two language models, the Show-Attend-Tell and the GPT-3, to generate comprehensive and descriptive radiology records. The proposed combination of these models generates a textual summary with the essential information about pathologies found, their location, and the 2D heatmaps localizing each pathology on the original X-Ray scans. The proposed model is tested on two medical datasets, the Open-I, MIMIC-CXR, and the general-purpose MS-COCO. The results measured with the natural language assessment metrics prove their efficient applicability to the chest X-Ray image captioning.
\end{abstract}

\begin{document}

\flushbottom
\maketitle

\thispagestyle{empty}

\section{Introduction}
\label{sec:intro}
Medical imaging is indispensable in the current diagnostic workflows. Out of the plethora of existing imaging modalities, X-Ray remains one of the most widely-used visualization methods in many hospitals around the world, because it is inexpensive and easily accessible \cite{irvin2019chexpert}.
Analyzing and interpreting X-ray images is especially crucial for diagnosing and monitoring a wide range of lung diseases, including pneumonia \cite{DBLP:journals/jamia/Demner-FushmanK16}, pneumothorax\cite{Chan_2018}, and COVID-19 complications \cite{maghdid2021diagnosing}. 

Today, the generation of a free-text description based on clinical radiography results has become a convenient tool in clinical practice \cite{Monshi2020}. Having to study approximately 100 X-Rays daily\cite{Monshi2020}, radiologists are overloaded by the necessity to report their observations in writing, a tedious and time-consuming task that requires a deep domain-specific knowledge. 
The typical manual annotation overload can lead to several problems, such as missed findings, inconsistent quantification, and delay of a patient’s stay in the hospital, which brings increased costs for the treatment. Among all, the qualification of radiologists as far as the correct diagnosis establishing should be stated as major problems. 

In the COVID-19 era, there is a higher need for robust image captioning \cite{Monshi2020} framework. Thus, many healthcare systems outsource the medical image analysis task. Automatic generation of chest X-Ray medical reports using deep learning can assist and accelerate the diagnosis establishing  process followed by clinicians. Providing automated support for this task has the potential to ease clinical workflows and improve both care quality and standardization. We propose to apply a model that works perfectly on non-medical data, to the medical data.

\subsection{Medical background} \label{sec:med_background}
Radiology is the medical discipline that uses medical imaging to diagnose and treat diseases. Today, radiology actively implements new artificial intelligence approaches\cite{Gurgitano2021}. There are three types of radiologists - diagnostic radiologists, interventional radiologists and radiation oncologists. They all use medical imaging procedures such as X-Rays, computed tomography (CT), magnetic resonance imaging (MRI), nuclear medicine, positron emission tomography (PET) and ultrasound. Diagnostic radiologists interpret and report on images resulted from imaging procedures, diagnose the cause of patient's symptoms, recommend treatment and offer additional clinical tests. They specialize on different parts of human body - breast imaging (mammograms), cardiovascular radiology (heart and circulatory system), chest radiology (heart and lungs), gastrointestinal radiology (stomach, intestines and abdomen), etc. Interventional radiologists use radiology images to perform clinical procedures with minimally invasive techniques. They are often involved in treating cancers or tumors, heart diseases, strokes, blockages in the arteries and veins, fibroids in the uterus, back pains, liver and kidney problems. Radiation oncologists use radiation therapy to treat cancer.

\subsection{Technical background} \label{sec:tech_background}
Since image captioning is a multimodal problem, it draws a significant attention of both computer vision and natural language processing communities.
The latest surveys in the medical image captioning task \cite{Monshi2020,pavlopoulos-etal-2019-survey} offer a detailed description of domain knowledge from radiology and deep learning. The first architectures to address this problem were CNN-RNN models from \cite{vinyals2015show} and \cite{shin2016learning}. However, the latter show satisfactory results only on the single-pathology tasks. 

With the new concept of attention approach\cite{bahdanau2016neural}, more papers began to use visual attention \cite{xushow}, \cite{donahuelongterm} and \cite{zhang2017mdnet}, being the first to use attention on medical images. Authors of \cite{xushow} presented the model that can fix its attention on salient objects while generating the corresponding words in the output sequence. Shortly after the visual-attention concept was exposed, text-attention was introduced by authors of \cite{You2016, Wang2018, jing-etal-2018-automatic}. They used both semantic and visual attention, that allowed them to get high natural language generation (NLG) metrics on medical datasets. Authors of \cite{Wang2018} introduced a framework generating natural reports for the Chest-Xray14 dataset \cite{Wang2017} - TieNet. It was trained for solving several tasks such as classification, localization, and text generation.  It used a non-hierarchical CNN-LSTM approach together with the attention to semantic and visual features, as it allowed to beat the current state-of-the-art results. In the \cite{Gale2018}, bone fracture X-Ray reports were generated by identifying image features and filling text templates. Authors of \cite{jing-etal-2018-automatic} suggested a multi-task framework, that can both predict tags and generate texts using co-attention mechanism. This model is still not sufficient for producing accurate diagnosis from X-Rays as the produced texts still contained repeated sentences due to a lack of contextual coherence in the hierarchical models.
The authors of  \cite{yuan2019automatic} took advantage of a sentence-level attention mechanism in a late fusion fashion. They took advantage of the multi-view images using both frontal and lateral views from the Open-I dataset \cite{demner2016preparing}.

Authors of \cite{Zhang2020} proposed to utilize a pre-constructed knowledge graph embedding module (extracted from the Open-I images using Chexnet models \cite{rajpurkar2017chexnet}) on multiple disease findings to assist the report generation process. Authors of \cite{tuluptceva2020anomaly} exposed an anomaly detection method for detecting abnormalities on chest X-Rays with deep perceptual autoencoders. The authors of  \cite{liu2019clinically} first generated topics for sentences using reinforcement learning (RL) followed by the word decoder sequence generation from the topic with attention to the original images. RL was used for tuning to optimize readability. We solve this problem in a simpler method without losing in quality. To extract topics, we use the NegBio labeller \cite{articleneg, Wang2017}, which provides topics from clinical reports. We add these topics to the beginning of the medical report, for our model to understand where exactly the text should be generated.

The paper in \cite{ni-etal-2020-learning} dives into reporting abnormal findings on radiology images. The proposed method learns conditional visual-semantic embeddings in radiology images and reports further used to measure the similarity between image regions and medical reports. This by optimizing a triplet ranking loss. The authors of \cite{syedamahmood2020chest} developed an algorithm that learns fine-grained description of findings from images and uses their pattern of occurrences to retrieve and customize similar reports from a large report database. The work in \cite{Liu_Jingyu_2019} proposed a Contrast Induced Attention Network (CIA-Net), using contrastive learning on the aligned positive and negative samples for the disease localization on the chest X-Ray images. The work in \cite{pmlr-v121-cohen20a} studies the cross-domain performance, agreement between models, and model representations for X-Rays diagnostic prediction tasks. The authors test for concept similarity by regularizing a network to group tasks across multiple datasets together and observe variation across the tasks. The model in \cite{Rodin2019} generates a short textual summary with essential information on the found pathologies along with their location and severity. The model is trained on only 2\% of the MIMIC-CXR dataset, and generates short reports. Although, in this work, we train on whole MIMIC-CXR and generate full-text report.

Authors of \cite{DevlinCLT19, ziegler2020encoderagnostic, ALFARGHALY2021100557, chen2020generating, Xiong2019} attempted to use transformer-based models as decoders in the image captioning domain. The  \cite{chen2020generating} affirmed affirmed to generate radiology reports through the custom transformer with additional memory-driven unit. Another model was introduced in \cite{Xiong2019} where encoder detects regions of interest via a bottom-up attention module and extracts top-down visual features. In this study, the decoder is presented as a custom transformer. For example, the paper in \cite{ziegler2020encoderagnostic} proposes an approach called ''pseudo self-attention''. Its main idea is to incorporate the conditioning input as a pseudo history to a pretrained transformer. They add a new key and value weights in the self-attention module to be projected onto the decoder’s self-attention space, while \cite{ALFARGHALY2021100557} focuses on visual and weighted semantic features. 
\subsection{Contributions} \label{sec:contributions}
In the current paper, we address all the problems mentioned above. The contributions of this paper are the following:
\begin{itemize}
    \item We introduce the new architecture for image captioning, based on combination of two language models with image-attention (SAT) and text-attention (GPT-3), which outperforming  current state-of-the-art models
    \item We introduce the new preprocessing pipeline for radiology reports, that allows to get higher NLG metrics
    \item We perform extensive experiments to show the effectiveness of the proposed methods
    \item Finally, we contribute into deep learning community with two language models trained on large MIMIC-CXR dataset
\end{itemize}

The rest of the paper is organized as follows: \autoref{sec:methods} describes two language models architecture separately, \autoref{sec:proposed_approach} provides the description of the proposed approach, \autoref{sec:experiments} describes datasets used and computing power utilized, \autoref{sec:results} and \autoref{sec:discussion} present, compare the results, while section 6 introduces the results and conclusions of the paper.
\section{Methods}
\label{sec:methods}
\subsection{Show Attend and Tell}
Show Attend and Tell (SAT) \cite{xushow} is an attention-based image caption generation neural net. Attention-based technique allows to get well interpretable results, which can be utilized by radiologist to ensure their findings on X-Ray. Including attention, the module gives the advantage to visualize where exactly  the model 'sees' the specific pathology. SAT consists of three blocks: Encoder, Attention module and Decoder. It takes an image, encodes it, attends each part of the image, and generates a $L$-length caption $\vz$, an encoded sequence of words from $W$-length vocabulary: 
\begin{equation}
\label{eq:caption}
    \vz = \left\{\vz_1, \ldots, \vz_{L} \right\},\mbox{ } \vz_i \in \RR^{W_{SAT}}
\end{equation}
\subsubsection{Encoder}
Encoder is a convolutional neural network (CNN). It encodes an image and outputs a set of $C$ vectors, each of which is a $D$-dimensional representation of the image corresponding part: 
\begin{equation}
    \label{eq:annotations}
    \va = \left\{\va_1, \ldots, \va_C \right\},\mbox{ } \va_i \in \RR^{D \times D}
\end{equation}
Here $C$ represents the number of channels in the output of the encoder. It  depends on the used type of the encoder: 1024 for DenseNet-121~\cite{8099726}, 512 for VGG-16\cite{DBLP:journals/corr/SimonyanZ14a}, 2048 for InceptionV3~\cite{DBLP:journals/corr/SzegedyVISW15} and ResNet-101~\cite{inproceedings2}. $D$ is a configurable parameter representing the encoded vectors size. Features are extracted from the lower convolutional layer prior to the fully connected layers, and are being passed through the Adaptive Average Pooling layer.  This allows the decoder to selectively focus on certain parts of an image by selecting a subset of all the feature vectors.

\subsubsection{Decoder with attention module}
\begin{figure}[htb!]
\centering \includegraphics[width=\linewidth]{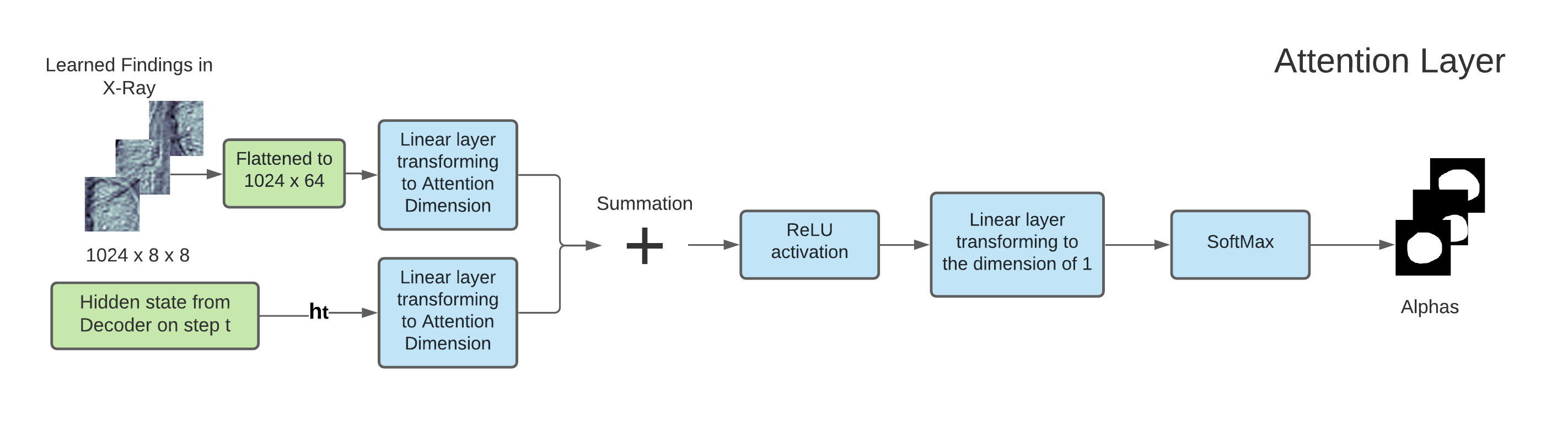}
\caption{Attention module used in SAT}
\label{fig:attention}
\end{figure}

The decoder is implemented as a LSTM neural network~\cite{hochreiter1997long}. It produces a caption by generating one word at every time step conditioned by the attention (context) vector, the previous hidden state and the previously generated words. The LSTM can be represented as the following set of equations:
\begin{align}
\label{eq:lstm_gates}
\begin{pmatrix}
\vi_t \\
\vf_t \\ 
\vo_t \\
\vg_t \\
\end{pmatrix} =
&
\begin{pmatrix}
\sigmoid \\
\sigmoid \\ 
\sigmoid \\
\tanh \\
\end{pmatrix}
T_{D+m+n, n}
\begin{pmatrix}
\vE\vz_{t-1}\\
\vh_{t-1}\\
\hat{\va_t}\\
\end{pmatrix}
\\
\label{eq:lstm_memory}
\vc_t &= \vf_t \odot \vc_{t-1} + \vi_t \odot \vg_t \\
\label{eq:lstm_hidden}
\vh_t &= \vo_t \odot \tanh (\vc_{t}). 
\end{align} 
Vectors $\vi_t$, $\vf_t$, $\vc_t$, $\vo_t$, $\vh_t$ represent the input/update gate activation vector, forgetting gate activation vector, memory or cell state vector, while outputting gate activation vector and hidden state of the LSTM respectively. $T_{s,t}$ is an affine transformation, such that $\RR^{s} \rightarrow \RR^{t}$ with non-zero bias. $m$ denotes the embedding dimension, while $n$ represents LSTM dimension. $\sigma$ and $\odot$ stand for the sigmoid activation function and element-wise multiplication, respectively. $\vE\in\RR^{m\times L}$ is an embedding matrix. The vector $\hat{\va} \in \RR^{D}$ holds the visual information from a particular input location of the image at time $t$. Thus, $\hat{\va}$ called context vector. \newline Attention is a function $\phi$, that computes context vector $\hat{\va}_t$ from the encoded vectors $\va_i$ (\ref{eq:annotations}), produced by the encoder. The attention module generates a positive number $\alpha_i$ for each location $i$ on the image. This number can be interpreted as the relative importance to give to the location $i$, among others. Attention module realized as a multi-layer perceptron (MLP) with a softmax activation function, conditioned at the previous hidden state $h_{t-1}$ (\ref{eq:lstm_hidden}) of the LSTM. The attention module is depicted in \autoref{fig:attention}. Set of linear layers in MLP is denoted as a function $f_{\mbox{att}}$. The weights $\alpha_{ti}$ are computed with the help of the following equations: 
\begin{align}
    e_{ti} =& f_{\mbox{att}} (\va_i, \vh_{t-1}) \\
    \label{eq:alpha}
    \alpha_{ti} =& \frac{\exp(e_{ti})}{\sum_{p=1}^C \exp(e_{tp})}
\end{align}

The sum of weights $\alpha_{ti}$ (\ref{eq:alpha}) should be equal to 1 $\sum_{i = 1}^{C} \alpha_{ti} = 1$. The context vector $\hat{a}_t$ is computed by the \emph{attention function} $\phi$ with the set of encoded vectors $\va$ (\ref{eq:annotations}) and their corresponding weights $\alpha_{ti}$ (\ref{eq:alpha}) as inputs: $\hat{\va}_t = \phi\left( \left\{ \va_i \right\}, \left\{ \alpha_{ti} \right\} \right)$.
According to the original paper function, $\phi$ can be either 'soft' or 'hard' attention. Due to specific task of medical image caption, function $\phi$ was chosen to be the 'soft' attention, as it allows model to focus more on some specific parts of X-Rays from others and to detect pathologies and major organs such as heart, lung etc. It is named as a 'deterministic soft attention' and recognized as a weighted sum : $\phi\left( \left\{ \va_i \right\},
\left\{ \alpha_{ti} \right\}\right) = \sum_i^C \alpha_i \va_i$. Hence, context vector can be computed as:
\begin{align}
\label{eq:context2}
    \hat{\va}_t = \sum_i^C \alpha_i \va_{ti}
\end{align}

The initial memory state and hidden state of the LSTM are initialized with two separate multi-layer perceptrons ($\text{init-c}$ and $\text{init-h}$) with the encoded vectors $\va_i$ (\ref{eq:annotations}) for a faster convergence:
\begin{align} 
    \vc_0 = f_{\text{init-c}} (\frac{1}{C} \sum_i^C \va_i) \\
    \vh_0 = f_{\text{init-h}} (\frac{1}{C} \sum_i^C \va_i)
\end{align}

To compute the output of LSTM representing a probabilities vector the next word, a 'deep output layer' \cite{hochreiter1997long} was used. It looks both on the LSTM state $\vh_t$ (\ref{eq:lstm_hidden}), on context vector $\hat{\va}_t$ (\ref{eq:context2}) and the one previous word $\vz_{t-1}$ (\ref{eq:annotations}):
\begin{equation}
\label{eq:p-out}
P(\vz_t | \hat{\va}_t, \vz_{t-1}) = softmax(\vL_o(\vL_h\vh_t+ \vL_a \hat{\va}_t + \vE\vz_{t-1}))
\end{equation}
where $\vL_o\in\RR^{W\times m}$, $\vL_h\in\RR^{m\times n}$, $\vL_a\in\RR^{m\times D}$, and $\vE\in\RR^{m\times L}$ represent the embedding matrix. 

The authors in \cite{xushow} suggest to use the 'doubly stochastic attention', where $\sum_t \alpha_{ti} \approx 1$. This can be interpreted as encouraging the model to pay equal attention to every part of the image. Yet, this method is not relevant for X-Rays, as each part of the chest is almost at the same position from image to image. If the model learned, \textit{e.g.}, that heart is in its specific position, a model doesn't have to search for the heart somewhere else. 
The model is trained in an end-to-end manner by minimizing the cross-entropy loss $L_{CE}$ between vector with a softmaxed distribution probability of next word and true caption as
$L_{CE} =  -\log(P(\textbf{z}|\textbf{a}))$.

\subsection{Generative Pretrained Transformer}
Generative Pretrained Transformer (GPT-3)\cite{brown2020language} is a large transformer-based language model with $1.75 \times 10^{11}$ parameters, trained on $570$ GB of text. GPT-3 can be used to generate realistic continuations texts from the arbitrary domain.
Basically, GPT-3 is a transformer that can look at a part of the sentence and predict the next word, thus being a language model. The original transformer~\cite{NIPS2017} is made up of encoder stack and decoder stack, in which encoders and decoders stacked upon each other. Whereas GPT-3 is built using just decoder blocks. One decoder block consists of Masked Self-Attention layer and Feed-Forward neural network. It is called Masked as it pays attention only to previous inputs. The input should be encoded prior to going into the decoder block.  In transformers and in the GPT-3 particularly, there are two subsequent encodings: Byte Pair Token Encoding and Positional Encoding. Byte Pair Encoding (BPE) is a simple data compression technique that iteratively replaces the most frequent pair of bytes in a sequence with a single, unused byte. The algorithm compresses data by finding the most frequently occurring pairs of adjacent subtokens in the data and replacing all instances of the pair with a single subword. The algorithm repeats this process until no further compression is possible. Such tokenization avoids adding a special \texttt{<unk>} token to the vocabulary, as now all words can be encoded and obtained by combination of subwords from the vocabulary.

\section{Proposed Architecture}
\label{sec:proposed_approach}
We introduce two architectures for X-Ray image captioning. The overall goal of our approach is to improve the quality of Encoder-Decoder generated clinical records by using the GPT-3 language model. The suggested model consists of two parts: the Encoder, Decoder (LSTM) with an attention module and the GPT-3. While the Encoder with LSTM detects pathologies and indicates zones of higher attention demand, the GPT-3 takes it as input and writes a comprehensive medical report.

\begin{figure}[htb!]
\centering \includegraphics[width=\linewidth]{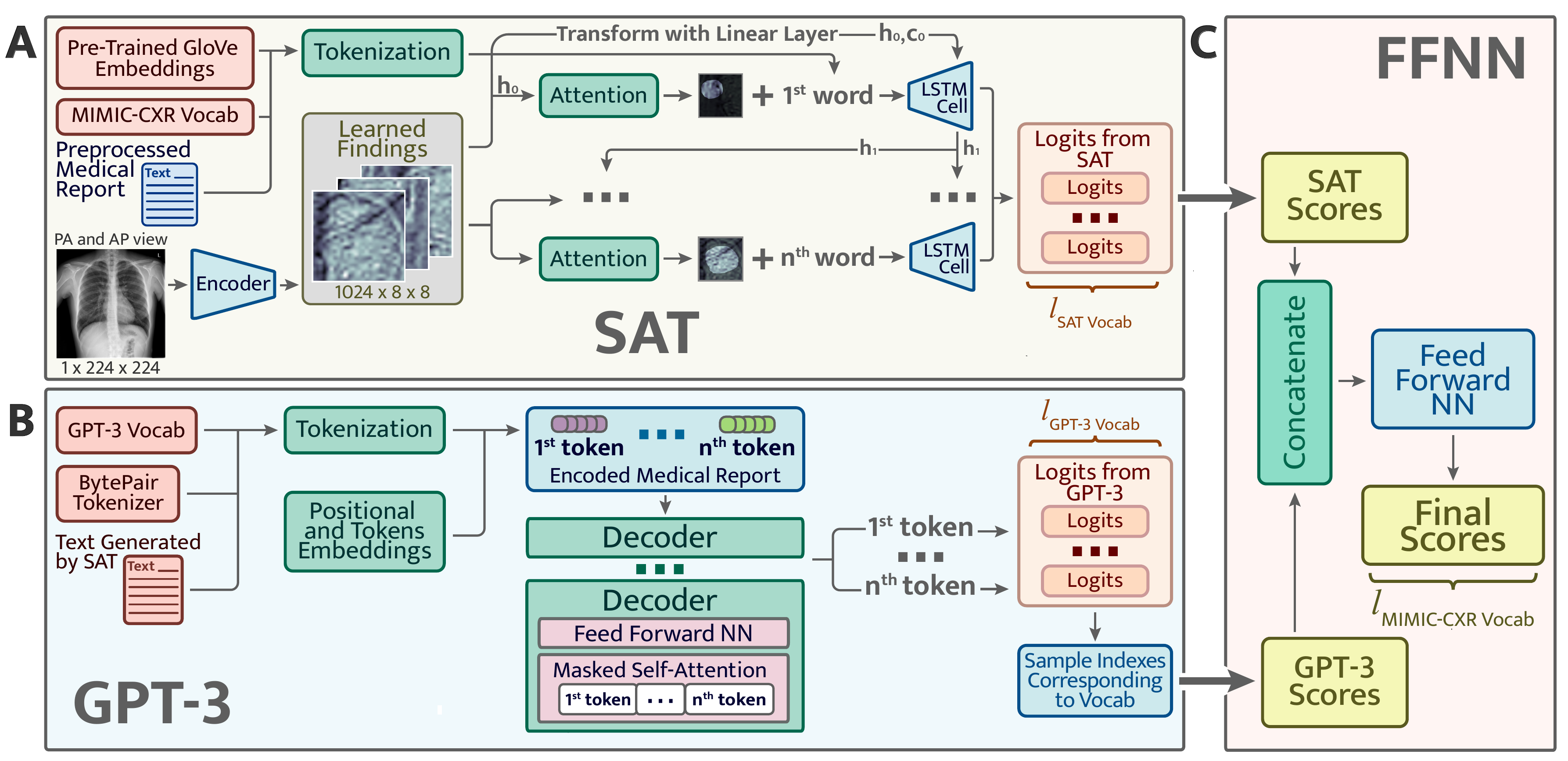}
\caption{The first approach. Learn the joint distribution of two models. The drawback is in sampling from the GPT-3 distribution.}
\label{fig:jointdistr}
\end{figure}

There are two possible approaches for this task. The first one consists in forcing models to learn joint word distribution. Within this method (Fig. \ref{fig:jointdistr}), both models \textbf{A} and \textbf{B} output scores for the next word in a sentence. Afterwards, due to concatenating these scores and pushing them through the feed-forward neural net  \textbf{C}, we get the final scores for upcoming word. Whilst the disadvantage of this approach is the following: the GPT-3 model has its own vocabulary built by the Byte Pair Tokenizer. This vocabulary is different from the one used by the Show Attend and Tell. We need to take from continuous GPT-3 distribution separate scores corresponding to the words present in the Show Attend and Tell vocabulary. This turns continuous distribution from the GPT-3 into discrete and hence, while we don't use all the potential generation power from the GPT-3. 

\begin{figure}[htb!]
\centering \includegraphics[width=\linewidth]{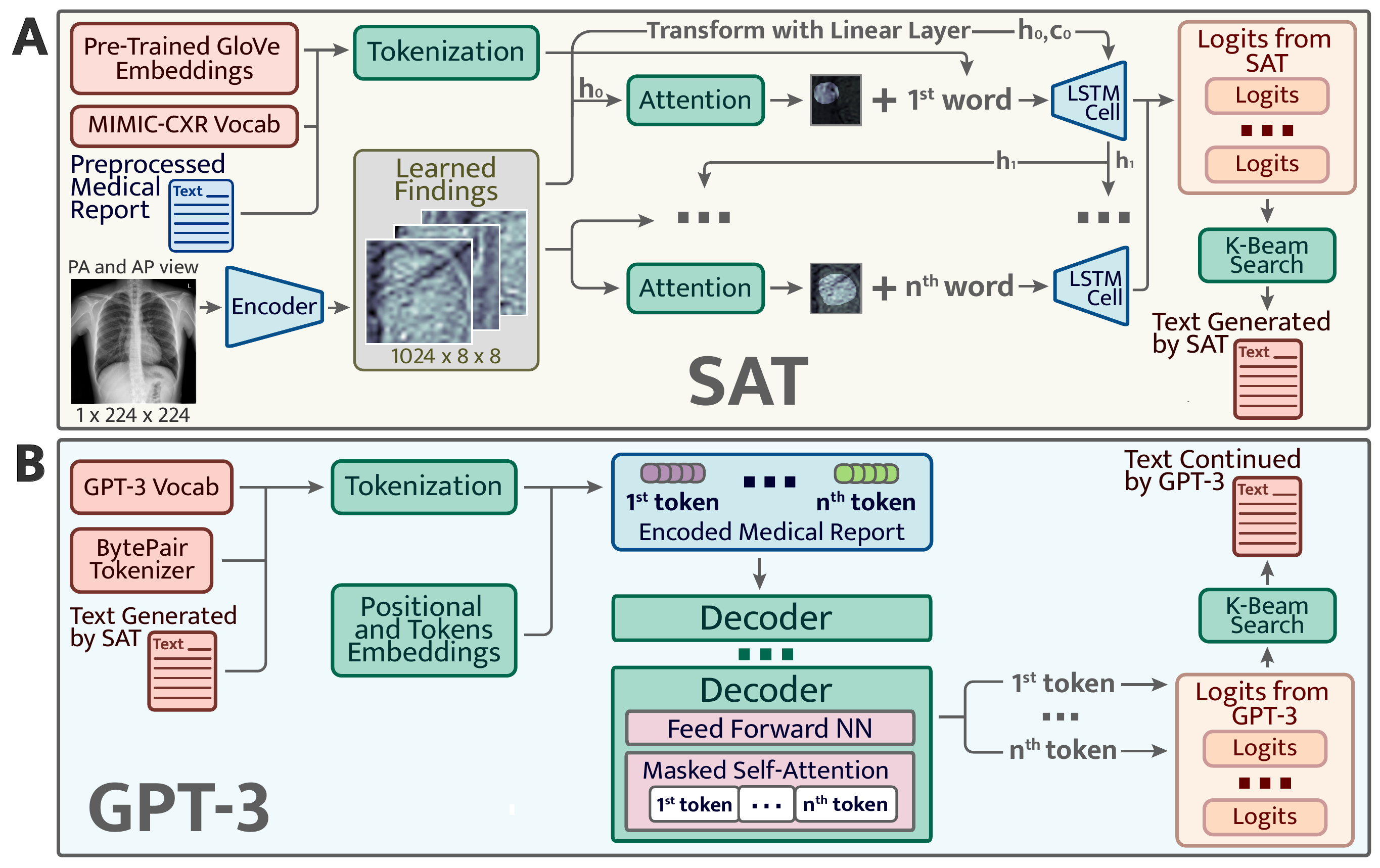}
\caption{Second approach. Pretrained GPT-3 (\textbf{B}) continues text generated by SAT (\textbf{A}).}
\label{fig:continue}
\end{figure}

The second method shown in Fig. \ref{fig:continue} consists in fine-tuning both models on the MIMIC-CXR dataset and using them one after another. Show Attend and Tell \textbf{A} gets an image as an input and generates a report based on the data found on X-Ray with an Attention module. It learns where to focus and gives a seed for the GPT-3 \textbf{B} to continue generating text. The GPT-3 was fine-tuned on MIMIC-CXR in self-supervised manner using the  Huggingface framework \cite{wolf-etal-2020-transformers}. It learns to predict the next word in the text. The GPT-3 continues the report outputed by SAT and generates a detailed and complete clinical report based on pathologies found by SAT. Such an approach is better for the GPT-3 as it gets more context as input (from SAT) than in the first approach. Thus, the second approach performs better, and was hence chosen by the authors of this paper as the main architecture.

\subsection{First Language Model}\label{s:first}
The first part of the suggested model is realized as the Show Attend and Tell model (SAT), the encoder, to encode the image and the LSTM for decoding into sequence.
The encoder encodes the input image with 3 or 1 color channels into a smaller image with 'learned' channels. The resulted encoded images can be interpreted as a summary representation of findings in the X-Ray (Eq. \ref{eq:annotations}). Those encoders pretrained on the ImageNet \cite{deng2009imagenet} are not suitable for the medical image caption task, as chest X-Rays doesn't have objects, figures from everyday life. Thus, the DenseNet-121 from \cite{Cohen2020xrv} pretrained on the MIMIC-CXR dataset was taken. It was trained for the classification task on 18 labels : Atelectasis, Consolidation, Infiltration, Pneumothorax, Edema, Emphysema, Fibrosis, Effusion, Pneumonia, Pleural Thickening, Cardiomegaly, Nodule, Mass, Hernia, Lung Lesion, Fracture, Lung Opacity, and Enlarged Cardiomediastinum. Hence, the last classification layer was removed and features from the last convolutional layer were taken. These features were passed through the Adaptive Average Pooling layer. As a result, the image encoded parts were obtained.  They can be represented by the tensor with the following dimensions: ($ batch size \times C, D, D $) (Eq. \ref{eq:annotations}). $C$ stands for the number of channels or how many different image regions to consider. $D$ implies the dimension of the image encoded region. Furthermore, the fine-tune method for encoder was added. It enables or disables the calculation of gradients for the encoder's parameters through the last layers. Then, at every time step, the decoder with the attention module observes the encoded small images with findings and generates a caption word by word. The Encoder output is received and flattened to dimensions ($ batch size, C, D \times D $). Since captions are padded with special \texttt{<pad>} token, captions are sorted by decreasing lengths and at every time-step of generating a word, an effective batch size is computed in order not to process the \texttt{<pad>} token.

The Show Attend and Tell model was trained using the Teacher-Forcing method while at each step the input to the model was the ground truth word on this step and not the previous generated word. As a result, we can consider the SAT as a language model $\textbf{A}$. It gets a tokenized text of length $m$, an image as input and outputs a vector of probabilities for the next word at each time step $t$:

\begin{align}
\begin{split}
    \textbf{A} : \text{text,image} \rightarrow P_1(\vz^t | \text{true words} =  \vz^{<1>}\vz^{<2>}\dots\vz^{<t-1>}, \text{ image}), \\ t \in \{2, \dots m, \dots L\}, \\
    P_1 \in \RR^{m \times W_{SAT}}
\end{split}
\end{align}
where $W$ is the SAT vocabulary size and L is the length of generated report (Eq. \ref{eq:caption}). Where $P_1$ is computed as it is shown in the Eq. \ref{eq:p-out}. 


Over the training process the LSTM outputs a word with a maximum probability after the softmax layer. It is a greedy approach, yet there is also an option to use the K-Beam search. Authors of\cite{wiseman2016sequencetosequence} used the K-Beam during training, however this is not a common approach. In our experiments, the greedy approach was used within the training process, and we applied the K-Beam search over the inference stage. 

\subsection{Second Language Model}\label{s:second}
The second part of the architecture proposed is the GPT-3 being a language model. The GPT-3 is built from decoder blocks using the transformer architecture. At the same time, the decoder block consists of masked self-attention and feed-forward neural network (FFNN). The output yields the token probabilities, \textit{i.e.}, logits. The GPT-3 was pretrained separately on the MIMIC-CXR dataset and was then fine-tuned together with the SAT to enhance clinical reports.


We put a special token \texttt{<start>} at the end of the text generated by the SAT allowing the GPT-3 to understand where to start the generation process. We also used the K-Beam search after the GPT-3 generation and took the second best sentence from the output as a continuation. The pretrained GPT-3 performs as a separate language model $\textbf{B}$ and generates good records based on the input text or tags. The  GPT-3 generates report till the moment when it generates the special token \texttt{<|endoftext|>} token. We denote the length of the GPT-3 generated text as $l$

\begin{align}
\begin{split}
    \textbf{B} : \text{text} \rightarrow P_2(\vz^t | \text{true words} =  \vz^{<1>}\dots\vz^{<L>}<\vs>), \\ t \in \{L+1, \dots L + l\}, \\
\end{split}
\end{align}


\subsection{Combination of two language models}
We use a combination of two models placing them sequentially: the SAT model extracts visual features from the image and allows us to focus on its specific parts. The GPT-3 provides good and comprehensive text, based on what is found by the first model. Thus, the predictions from the first model improve those of the second language model.

\subsection{Evaluation metrics}
The common evaluation metrics used for image captioning are : bilingual evaluation understudy (\texttt{BLEU}) \cite{Papineni2002}, recall-oriented
understudy for gisting evaluation (\texttt{ROUGE}) \cite{Lin2004}, metric for evaluation of translation with explicit ordering (\texttt{METEOR}) \cite{banerjee-lavie-2005-meteor}, consensus-based image description evaluation (\texttt{CIDEr}) \cite{7299087}, and semantic propositional image caption evaluation (\texttt{SPICE}) \cite{spice2016}. The Microsoft Common Objects in Context \cite{chen2015microsoft} provides the kit with implementation of these metrics for the image caption task.

\section{Experiments}
\label{sec:experiments}
\subsection{Datasets}\label{sec:dataset}
For training and evaluation of medical image captioning, we use three publicly available datasets. Two of them are medical images datasets and the third one is a general-purpose one.

\paragraph{MIMIC-CXR}
The MIMIC Chest X-Ray (MIMIC-CXR)\cite{MIMIC-CXR} dataset is a large publicly available dataset of chest radiographs in DICOM format with free-text radiology reports. This dataset consists of 377,110 images corresponding to 227835 radiographic studies performed at the Beth Israel Deaconess Medical Center in Boston, MA. 

\paragraph{Open-I} The Indiana University Chest X-Ray Collection (IU X-Ray)\cite{demner2016preparing} contains radiology reports associated with X-Ray images. This dataset contains 7470 image-report pairs. All the reports enclose the following sections: impression, findings, tags, comparison, and indication. We use the concatenation of impression and findings as the target captions

\paragraph{MSCOCO}
Microsoft Common Objects in Context dataset (MS COCO dataset)\cite{MSCOCO} is large-scale non-medical dataset for scene understanding. The dataset is commonly used for training and benchmark object detection, segmentation, and captioning algorithms.

\subsection{Image preprocessing}
Hierarchical Data Format (HDF5) \cite{hdf5} dataset was used to store all images. X-Rays are in gray-scale and have one channel. To process them with the pre-trained CNN DenseNet-121, we used 1 channel image. Each image was resized to the size of 224$\times$224 pixels, normalized to the range from 0 to 1, and converted to the \texttt{float32} type and stored in the HDF5 dataset.

\subsection{Image captions pre-processing}
Following the logic in\cite{Jing2018}, a medical report is considered as a concatenation of Impression and Findings sections, if both of these sections is empty then this report was excluded. This resulted in 360666 DICOMs with reports for MIMIC-CXR dataset. The text records are pre-processed by converting all tokens to lowercase, removing all non-alphanumerical tokens. For our experiments we used 75\% of data for training,  24,75 \% for validation and 0.25\% for testing.

The MIMIC-CXR database was used to access metadata and labels derived from free-text radiology reports. These labels were extracted using NegBio tool \cite{articleneg, Wang2017} that outputs one of 14 pathologies along with their severity and (or) absence. To generate more accurate reports, we added the extracted labels to the beginning of the report. This allows language models to know the summary of the report for a more precise description generation.

We additionally formed the abbreviations dictionary of 150$+$ words from the Unified Medical Language System (UMLS)\cite{Bodenreider2004}. We also extended our dictionary size with several commonly used medical terms from the Medical Concept Annotation Tool\cite{kraljevic2021multidomain}.

\subsection{Training of the Neural Network}
The pipeline is implemented using PyTorch. Experiments were conducted on a server running the Ubuntu 16.04 (32 GB RAM). All models were trained with NVIDIA Tesla V100 GPU (32 GB RAM). 
In all experiments, we use a 5-fold cross-validation and reported the mean performance. 
The SAT was trained for 70 epochs with batch size of 16, embedding dimension of 100, attention and decoder dimension of 512, dropout value 0.1. The encoder and decoder learning rates were $4\times 10^{-7}$ and $3\times 10^{-7}$, respectively. The Cross Entropy loss was used for training.
The best model is chosen according to the highest geometric mean of BLEU-n, as it is done in other works\cite{BLEU}. SAT was trained in Teacher-Forcing technique, while the Greedy approach is used for counting metrics. The GPT-3 small was fine-tuned with the MIMIC-CXR dataset for 30 epochs with batch size of 4, learning rate of $5\times 10^{-5}$, the Adam epsilon of $1\times 10^{-8}$, where the block size equals 1024, with clipping gradients, which are bigger than 1.0. It was fine-tuned in a self-supervised manner as a language model. No data augmentation was applied.

\section{Results \& Discussion}
\subsection{Quantitative results}
\label{sec:results}
The quantitative results for the baseline models, preceding works and our models are presented in Table \ref{tbl:metrics}. Models were evaluated on the most common Open-I dataset as well as on the big and rarely reported MIMIC-CXR dataset with free-text radiology reports. We implemented the most commonly used metrics for evaluation - \texttt{BLEU-n}, \texttt{CIDEr} and \texttt{ROUGE}. The proposed approach outperforms the existing models in terms of the NLG metrics -  \texttt{BLEU-n}, \texttt{CIDEr} and \texttt{ROUGE}. 

We additionally provided our model performance illustrations in Table \ref{tbl:samples} containing the original X-Ray images from the MIMIC-CXR dataset, the ground truth expert label and the model prediction (SAT + GPT-3). We manually underlined the similarities and identical diagnoses in texts to guide the eye.
\begin{table}[]
\centering
    \small
\resizebox{\textwidth}{!}{%
\begin{tabular}{ccllcccccc}
\toprule
\multicolumn{1}{l}{}                         & \multicolumn{3}{c}{Model}                & CIDEr & ROUGE\_L & BLEU-1 & BLEU-2 & BLEU-3 & BLEU-4 \\ \midrule
\multirow{6}{*}{{MIMIC-CXR}}                   & \multicolumn{3}{c}{S\&T \cite{vinyals2015show}}                  & 0.886 & 0.300    & 0.307  & 0.201  & 0.137  & 0.093  \\
                     & \multicolumn{3}{c}{Original SAT \cite{xushow}} &  0.967 &    0.288       & 0.318 & 0.205 & 0.137 &  0.093\\
                     & \multicolumn{3}{c}{TieNet \cite{Wang2018}} & 1.004 &  0.296         & 0.332 & 0.212 & 0.142 & 0.095 \\
                     & \multicolumn{3}{c}{NLG \cite{liu2019clinically}} & 1.153 &    0.307       & 0.352 & 0.223 & 0.153 & 0.104 \\ \cline{2-10}
                     & \multicolumn{3}{c}{\bf SAT} &  1.986 &   0.478         & 0.634 & 0.549 & 0.451 &  0.383\\
                     & \multicolumn{3}{c}{\textbf{SAT + GPT-3}} & \bf 1.989 &   \bf 0.480        & \bf 0.725 & \bf 0.626 & \bf 0.505 & \bf 0.418 \\ \midrule
\multirow{9}{*}{{Open-I}}                      & \multicolumn{3}{c}{Co-Attention \cite{Jing2018}} & 0.327 & 0.447    & 0.517  & 0.386  & \bf 0.306  & \bf 0.247  \\
                     & \multicolumn{3}{c}{TieNet \cite{Wang2018} } & - &    0.311       & 0.330 & 0.194 & 0.124 & 0.081 \\
                     & \multicolumn{3}{c}{CNN-RNN \cite{vinyals2015show}} & 0.111 & 0.267 & 0.316 & 0.211 & 0.140 & 0.095 \\
                     & \multicolumn{3}{c}{LRCN \cite{donahue2016longterm} } & 0.190 &  0.278         & 0.369 & 0.229 & 0.149 & 0.138 \\
                     & \multicolumn{3}{c}{ATT-RK \cite{You2016} } & 0.155 &    0.323       &  0.369 & 0.226 & 0.151 & 0.108 \\
                     & \multicolumn{3}{c}{CDGPT2 \cite{ALFARGHALY2021100557} } & 0.257 &         0.289  & 0.387 & 0.245 & 0.166 & 0.111 \\
                     & \multicolumn{3}{c}{Original SAT \cite{xushow} } & 0.320 &   0.361        & 0.433 & 0.281 & 0.194 & 0.138 \\ \cline{2-10}
                     & \multicolumn{3}{c}{\bf SAT} & 0.699 &    0.413       & 0.407 & 0.258 & 0.210 & 0.125 \\
                     & \multicolumn{3}{c}{\textbf{SAT + GPT-3}} & \bf 0.701 &  \bf 0.450         & \bf 0.520 & \bf 0.390 & 0.296 &  0.235\\ \midrule
\multicolumn{1}{l}{\multirow{4}{*}{{MS-COCO}}} & \multicolumn{3}{c}{BRNN \cite{karpathy2015deep}} & - &     -      & 0.642 &  0.451 & 0.304 &  0.203\\
\multicolumn{1}{l}{} & \multicolumn{3}{c}{Original SAT \cite{xushow}} & - &     -      & 0.718 &  0.504 & 0.357 &  0.250\\ \cline{2-10}
\multicolumn{1}{l}{} & \multicolumn{3}{c}{\bf SAT} & 1.300   &    0.592         &  0.815   &  0.663   & 0.516   &  0.395  \\ 
\multicolumn{1}{l}{} & \multicolumn{3}{c}{\bf SAT + GPT-3} & \bf 1.360  &    \bf 0.606        & \bf 0.821  & \bf 0.672  & \bf 0.529  &  \bf 0.409 \\ \bottomrule
\end{tabular}%
}
\caption{\small Reported mean performance using word-overlap metrics for two medical radiology datasets and one non-medical for general purpose. Here SAT stands for the model implemented by us and trained with the preprocessed MIMIC-CXR data. BLUE-n denotes the BLEU score that uses up to n-grams.}
\label{tbl:metrics}
\end{table}

\begin{table}[t]
\setlength{\tabcolsep}{0.6em}
\centering
\tiny
\sffamily
\resizebox{\textwidth}{!}{%
\begin{tabular}{L{3.8cm}	L{4.0cm}	L{4.5cm}}
    \multicolumn{1}{c}{\bf {\normalsize Chest X-Ray}}
  	& \multicolumn{1}{c}{\bf {\normalsize Ground Truth}}
  	& \multicolumn{1}{c}{\bf {\normalsize Our predictions}}\\ \cline{1-3}
 \includegraphics[width=0.6\linewidth]{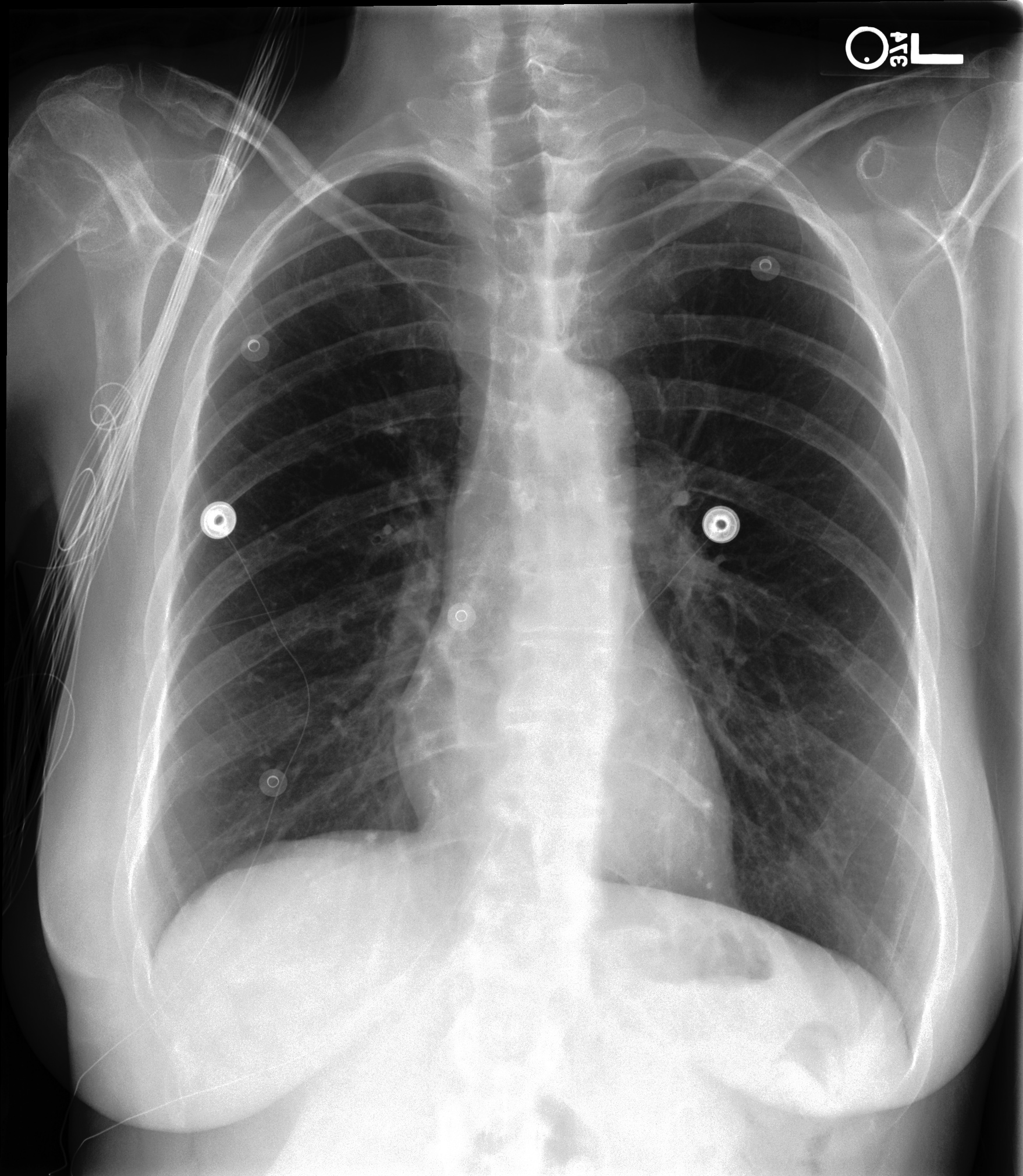}	& Lungs remain well inflated \ul{without evidence of focal airspace consolidation, pleural effusions, pulmonary edema or pneumothorax}.Irregularity in the right humeral neck is related to a known healing fracture secondary to recent fall. PA and lateral views of the chest \underline{\hspace{0.4cm}} at 09:55 are submitted. & \ul{no findings. no pneumonia. no pleural effusion. no edema. there is little change and no evidence of acute cardiopulmonary disease. no pneumonia, vascular congestion, pleural effusion}.of incidental note is an azygos fissure, of no clinical significance. this raises possibility of a normal variant. \\ \cline{1-3}
 \includegraphics[width=0.6\linewidth]{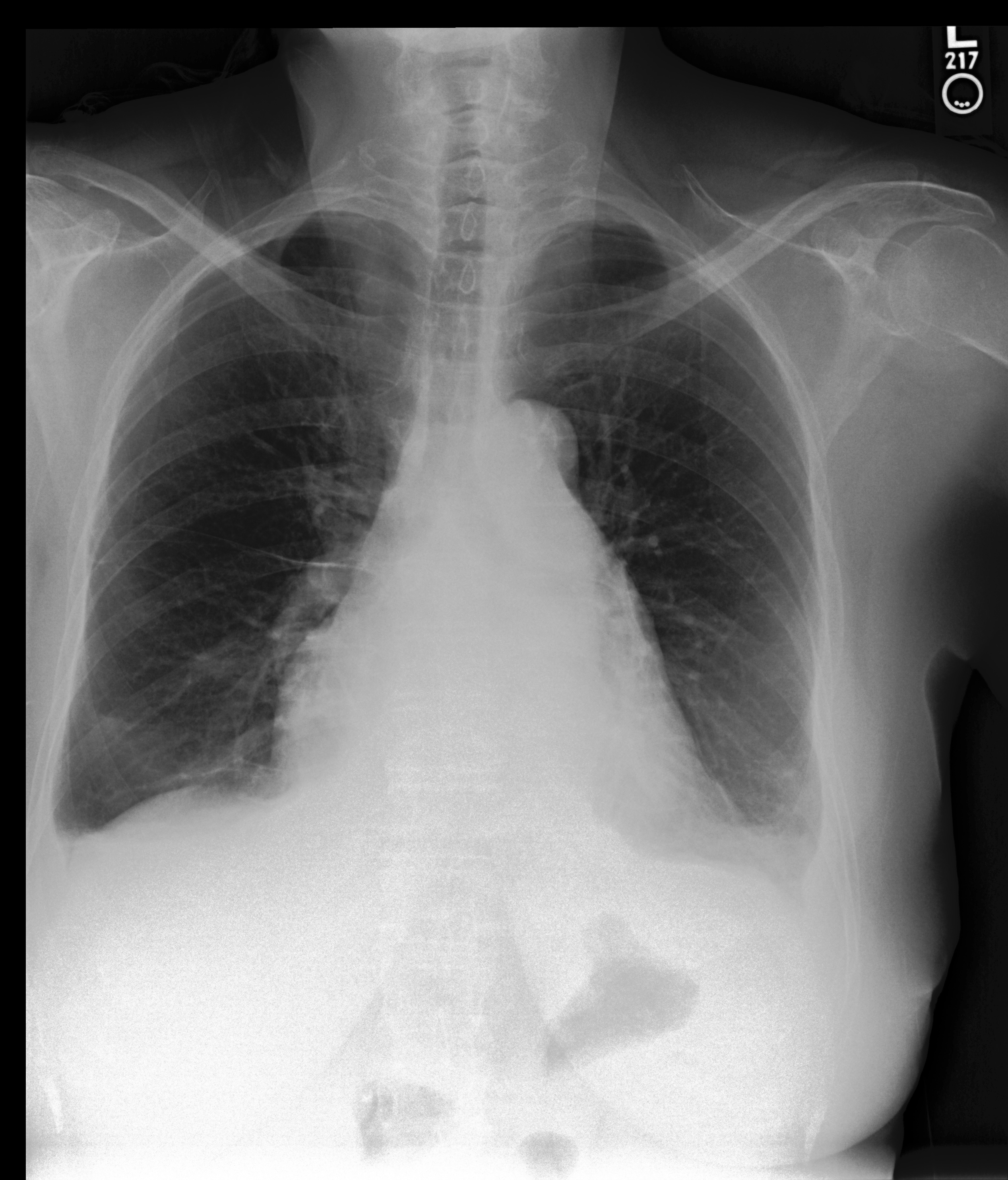}	& 1.  Stable \ul{bilateral small pleural effusions and atelectasis}. 2.  \ul{Enlarged pulmonary artery, suggesting pulmonary hypertension. No significant interval change.  Bilateral small pleural effusions and adjacent atelectasis are overall unchanged}.  The heart is top-normal in size, unchanged.  The pulmonary artery is enlarged, suggesting \ul{pulmonary hypertension}.  No demand, focal consolidation to suggest pneumonia, or pneumothorax. & \ul{pleural effusion present. lung opacity present. no edema. cardiomegaly present. atelectasis present.} as compared to previous radiograph, there is an increase in extent of a pre existing \ul{small left pleural effusion with subsequent atelectasis at left lung bases}. otherwise, radiograph is unchanged. moderate cardiomegaly. mild fluid overload no overt pulmonary edema. \ul{no new focal parenchymal opacities suggesting pneumonia}. unchanged position of right pectoral port a cath.\\ \cline{1-3}
 \includegraphics[width=0.6\linewidth]{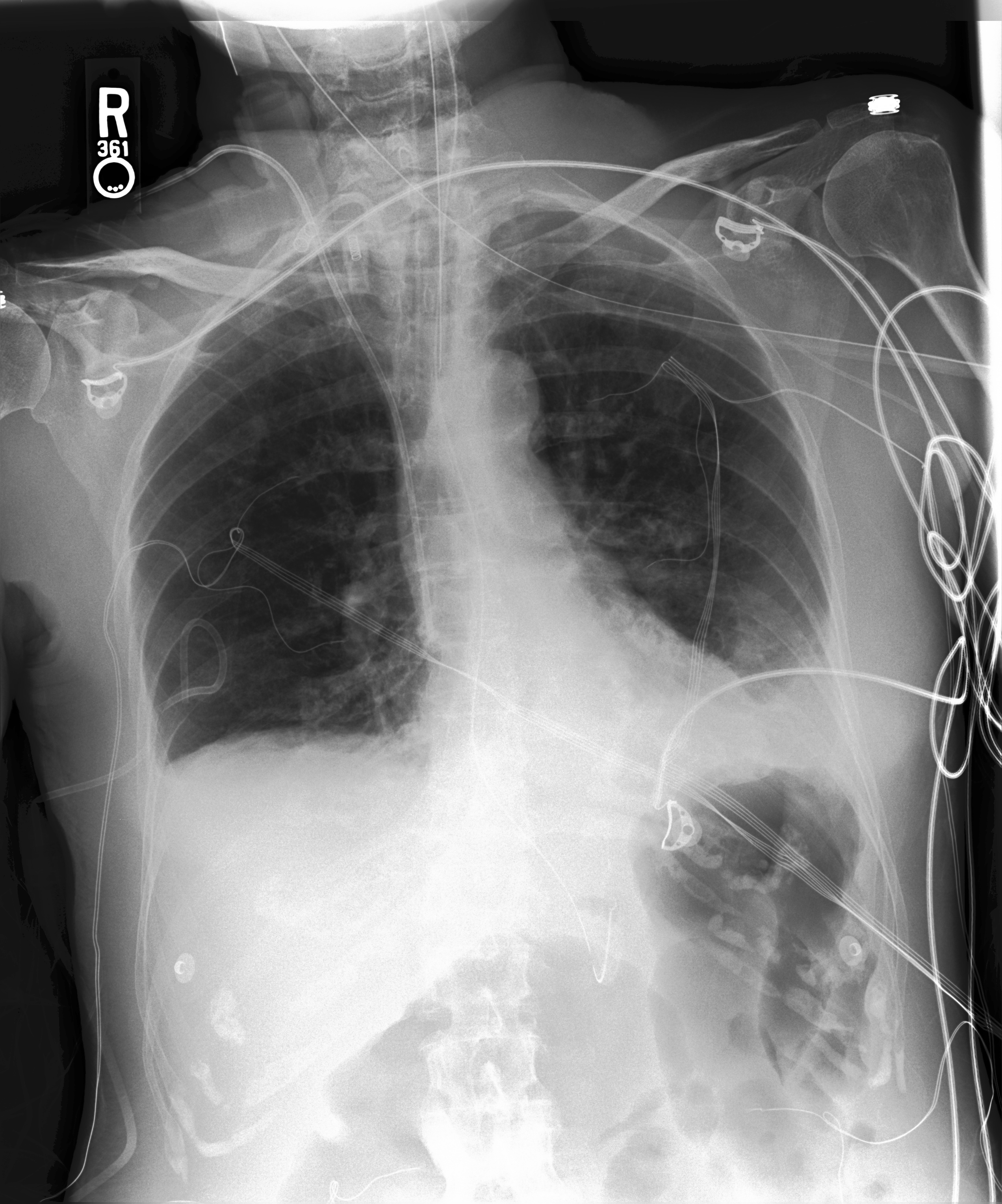}	& There is decrease in now small right \ul{pleural effusion}.  \ul{There is no pneumothorax}.  There is a new right pacer pigtail catheter. Cardiomediastinal contours are unchanged.  Lines and tubes are in standard position.  \ul{Left lower lobe opacities, a combination of pleural effusion and atelectasis}, are unchanged.& \ul{uncertain pneumonia. pleural effusion present. lung opacity present. atelectasis present. bilateral pleural effusions, left greater than right}. bibasilar opacities potentially atelectasis in setting of low lung volumes. infection be excluded. frontal and lateral views of chest demonstrate \ul{low lung volumes, which accentuate bronchovascular markings}. there are \ul{small bilateral pleural effusions, right greater than left, with adjacent atelectasis}. there is \ul{no focal consolidation pneumothorax}. cardiomediastinal silhouette is within normal limits. surgical clips are seen in right upper quadrant of abdomen. aortic arch calcifications are noted. \\ \cline{1-3}
 \includegraphics[width=0.6\linewidth]{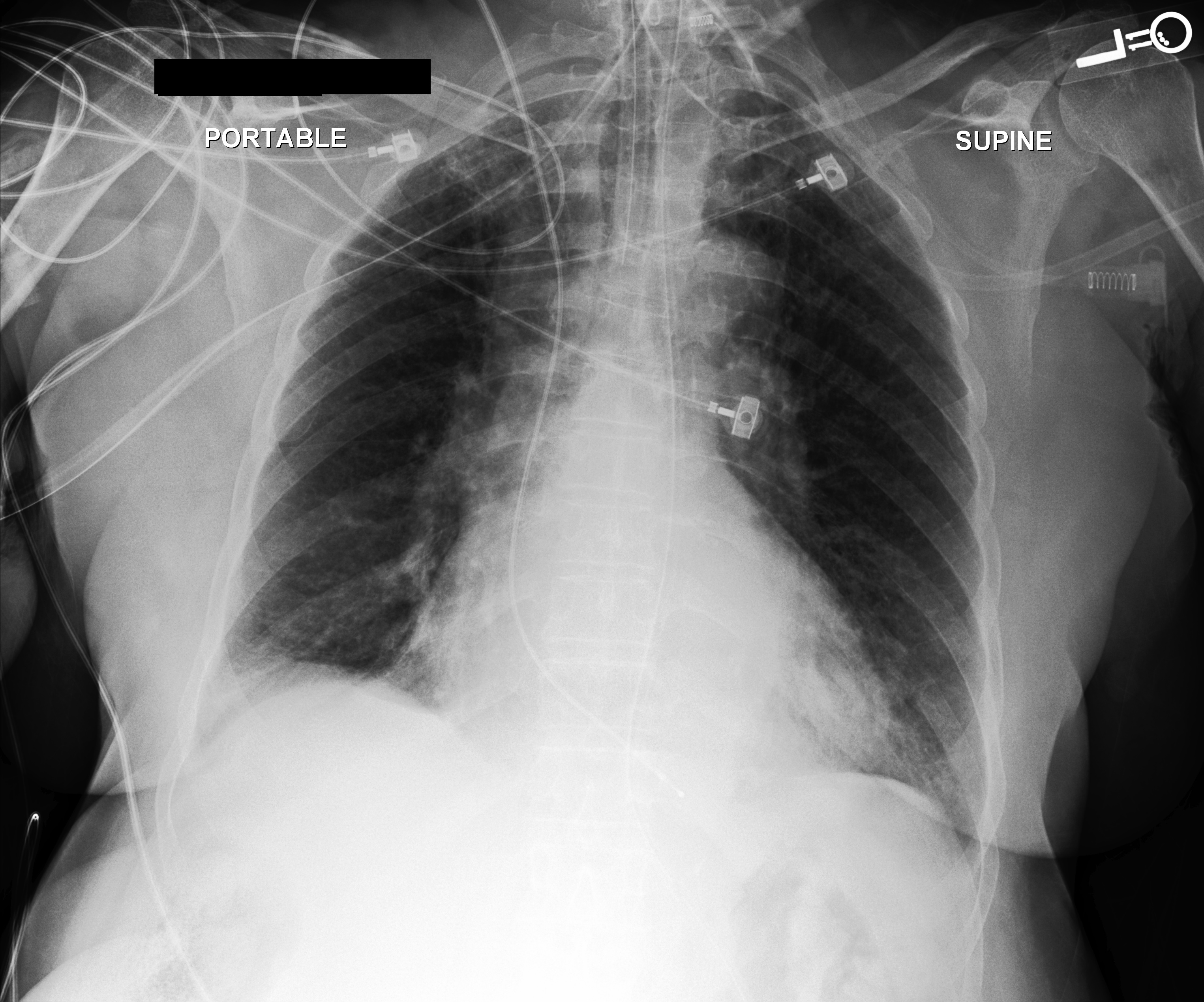}	& Compared to prior chest radiographs \underline{\hspace{0.4cm}} through \underline{\hspace{0.4cm}}. Previous mild \ul{pulmonary edema has improved, moderate cardiomegaly and mediastinal vascular engorgement have not.}  ET tube, right transjugular temporary pacer lead are in standard placements and an esophageal drainage tube passes into the stomach and out of view. \ul{Pleural effusions are presumed but not substantial.  No pneumothorax.}	& \ul{support devices present. no pneumothorax. pleural effusion present. lung opacity present. uncertain enlarged cardiomediastinum. no edema. atelectasis present.} right internal jugular central line has its tip in distal superior vena cava. \ul{overall cardiac and mediastinal contours are likely stable given patient rotation on current study}. lung volumes remain low with patchy opacities at both bases likely reflecting atelectasis. blunting of both costophrenic angles may reflect small effusions. \\

\end{tabular}}

\rmfamily
\caption{\small Results of generated reports by the proposed SAT + GPT-3 model.}
\label{tbl:samples}

\end{table}

\subsection{Discussion}
\label{sec:discussion}


The first language model (SAT) learned to generate short summary at the beginning of the report, based on findings from the X-Ray to provide the finding details. This offers text generation direction seed for the second model. Performed preprocessing of medical reports allowed to get these high metrics. We also address the biased data problem by applying domain-specific text preprocessing while using the NegBio labeller. In a radiology database, the data is unbalanced because abnormal cases are rarer than the normal ones. The NegBio labeller allowed us to get a not negative-biased diagnosis clinical records as it added short sentences at the beginning of ground truth report, making this task closer (in some ways) to classification task, when the state-of-the-art models had already managed to achieve strong performance. The SAT also provides 2D heatmaps of pathologies localization, assisting and accelerating the diagnosis process followed by clinicians.

The second language model, the Generative Pretrained Transformer (GPT-3), showed promising results in the medical domain. It successfully continued texts from the first language model, taking into consideration all the findings provided. As GPT-3 is a large and smart transformer, it summarizes and provides more details on findings. Natural language generation metrics suggest using two language models subsequently. Such an approach can be considered as strong for the text generation.

The SAT followed by the GPT-3 outperformed the reported state-of-the-art (SOTA) models in all the 3 datasets considered. Notably, the proposed approach beats SOTA models on MIMIC-CXR demonstrating the highest performance in all the metrics measured. The performance for the main evaluation dataset, the Open-I, is also measured by the F1-score using micro-averaging and demonstrates 0.861 \textit{vs.} 0.840 for the proposed (SAT + GPT-3) model and the SAT, respectively.

Examples of the reports generated jointly via the Show-Attend-Tell + GPT-3 architecture, are shown in Table \ref{tbl:samples}. 
One may notice that some generated sentences are identical with the ground truth. For example, in both generated and true reports for the first X-Ray is ``no acute cardiopulmonary abnormality".
Some sentences close in their meaning, even, even if they are different in terms of chosen words and n-grams ("no pneumonia. no pleural effusion. no edema. ..." compared to `` without pulmonary edema or pneumothorax").

\section{Conclusions}
\label{sec:conclusion}

The authors of the current paper introduced a new technique of combining two language models for the medical image captioning task. Principally, the new preprocessing and squeezing approaches for clinical records were implemented along with a combined language model, where the first component is based on attention mechanism and the second represents a generative pretrained transformer. The proposed combination of models generates a descriptive textual summary with essential information on found pathologies along with their location and severity. Besides, the 2D heatmaps localize each pathology on the original X-Ray scans. The results measured with the natural language generation metrics on both the MIMIC-CXR and the Open-I datasets speak for an efficient applicability to the chest X-Ray image captioning task. This approach also provides well-interpretable results and allows to support medical decision making.

We investigated various approaches to the text from the angle of generation automatic X-Ray captioning. We proved that the Show-Attend-Tell is a strong baseline outperforming models with Transformer-based decoders. With the help of the GPT-3 pre-trained language model, we managed to improve this baseline. The simple method, whither the GPT-3 model finishes report started by the Show-Attend-Tell model, yields significant improvements of the standard text generation scores.

\section{Acknowledgements}
The authors of this paper thank Alexander Panchenko and Alexander Shvets for the helpful discussion.

\bibliographystyle{naturemag}
\bibliography{biblio}

\end{document}